\documentclass[twocolumn, switch]{article} 

\usepackage{preprint}

\usepackage{amsmath, amsthm, amssymb, amsfonts}
\usepackage{times}
\usepackage{float}
\usepackage{epsfig}
\usepackage{graphicx}
\usepackage{amsmath}
\usepackage{amssymb}
\usepackage{booktabs}
\usepackage{algorithm}
\usepackage{algpseudocode}
\usepackage{tikz}
\usepackage{enumitem}
\setlist[itemize]{noitemsep, nolistsep}
\usetikzlibrary{shapes.multipart, positioning}
\usepackage{graphicx}
\usepackage{subfig}
\usepackage{rotating}
\usepackage{array}

\usepackage[numbers,square]{natbib}
\usepackage[utf8]{inputenc}	
\usepackage[T1]{fontenc}	
\usepackage{xcolor}		
\usepackage[colorlinks = true,
            linkcolor = purple,
            urlcolor  = blue,
            citecolor = cyan,
            anchorcolor = black]{hyperref}	
            
\usepackage{booktabs} 		
\usepackage{nicefrac}		
\usepackage{microtype}		
\usepackage{lineno}		
\usepackage{float}			

\usepackage{lipsum}		
\usepackage{url} 
\usepackage{newfloat}
\DeclareFloatingEnvironment[name={Supplementary Figure}]{suppfigure}
\usepackage{sidecap}
\sidecaptionvpos{figure}{c}

\usepackage{titlesec}
\titlespacing\section{0pt}{12pt plus 3pt minus 3pt}{1pt plus 1pt minus 1pt}
\titlespacing\subsection{0pt}{10pt plus 3pt minus 3pt}{1pt plus 1pt minus 1pt}
\titlespacing\subsubsection{0pt}{8pt plus 3pt minus 3pt}{1pt plus 1pt minus 1pt}

\usepackage{tikz,xcolor,hyperref}

\definecolor{lime}{HTML}{A6CE39}

\title{Accelerated Fingerprint Enhancement: A GPU-Optimized Mixed Architecture Approach}

\usepackage{authblk}

\author[1]{André Brasil Vieira Wyzykowski}
\author[2]{Anil K. Jain}

\affil[1]{Department of Computer Science and Engineering, Michigan State University  {\tt\small \{wyzykow2,jain\}@msu.edu}}

\begin{document}

\twocolumn[ 
  \begin{@twocolumnfalse} 
  
\maketitle

\begin{abstract}
This document presents a \textbf{preliminary approach} to latent fingerprint enhancement, fundamentally designed around a mixed Unet architecture. It combines the capabilities of the Resnet-101 network and Unet encoder, aiming to form a potentially powerful composite. This combination, enhanced with attention mechanisms and forward skip connections, is intended to optimize the enhancement of ridge and minutiae features in fingerprints. One innovative element of this approach includes a novel Fingerprint Enhancement Gabor layer, specifically designed for GPU computations. This illustrates how modern computational resources might be harnessed to expedite enhancement. Given its potential functionality as either a CNN or Transformer layer, this Gabor layer could offer improved agility and processing speed to the system. However, it is important to note that this approach is still in the early stages of development and has not yet been fully validated through rigorous experiments. As such, it may \textbf{require additional time and testing} to establish its robustness and usability in the field of latent fingerprint enhancement. This includes improvements in processing speed, enhancement adaptability with distinct latent fingerprint types, and full validation in experimental approaches such as open-set (identification 1:N) and open-set validation, fingerprint quality evaluation, among others.
\end{abstract}
\vspace{0.12cm}

  \end{@twocolumnfalse} 
] 



\section{Introduction}
 \vspace{-0.34cm}

The objective of the forensic study is to enhance the detection and examination of latent fingerprints, which are often unseen traces on surfaces that constitute crucial evidence in criminal probes. Determining latent fingerprints can be arduous due to interference from the background, superimposed prints, and pollutants like dust, dirt, or grime. These elements can blur ridge configurations, complicating the distinction between authentic ridges and incidental marks. Furthermore, the process of gathering can inject additional disturbance and deteriorate the fingerprint quality.

\begin{figure}[h]
\centering
    \setlength{\tabcolsep}{1pt}
            \begin{tabular}{ccc}

            \footnotesize NIST SD27 \cite{garris2000nist} & \footnotesize MOLF \cite{sankaran2015multisensor} & \footnotesize MSP Latent \cite{yoon2015longitudinal}\\

            \includegraphics[height=2.4cm]{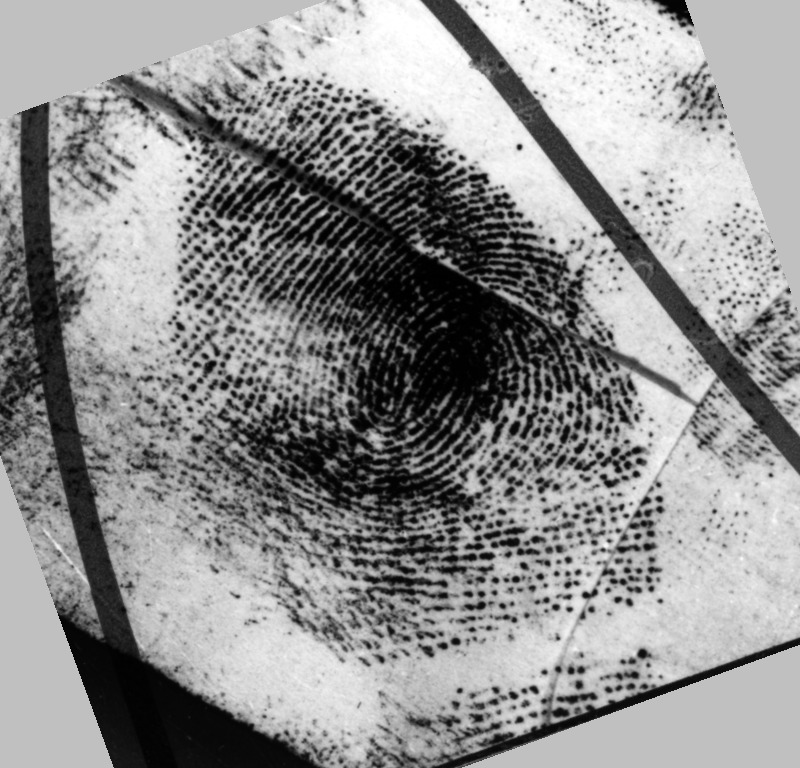}&
            \includegraphics[height=2.4cm]{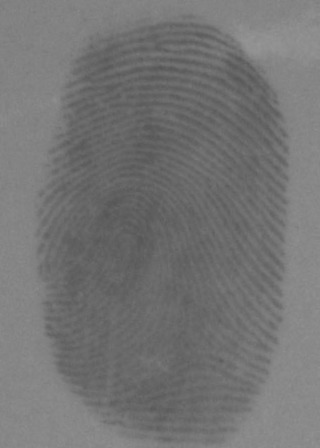}&
            \includegraphics[height=2.4cm]{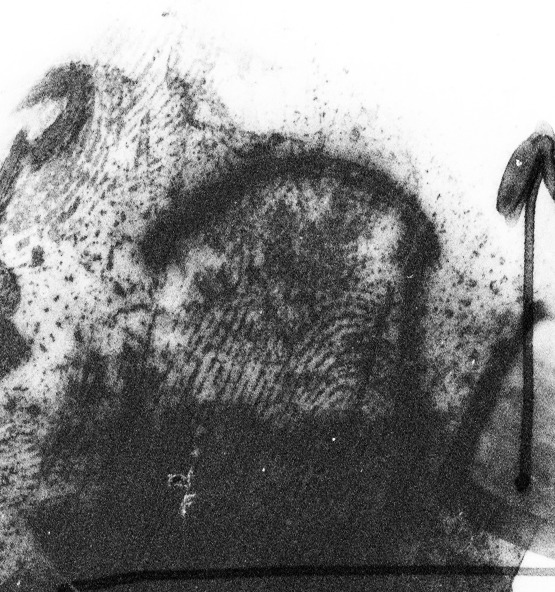}\\
            
            \includegraphics[height=2.4cm]{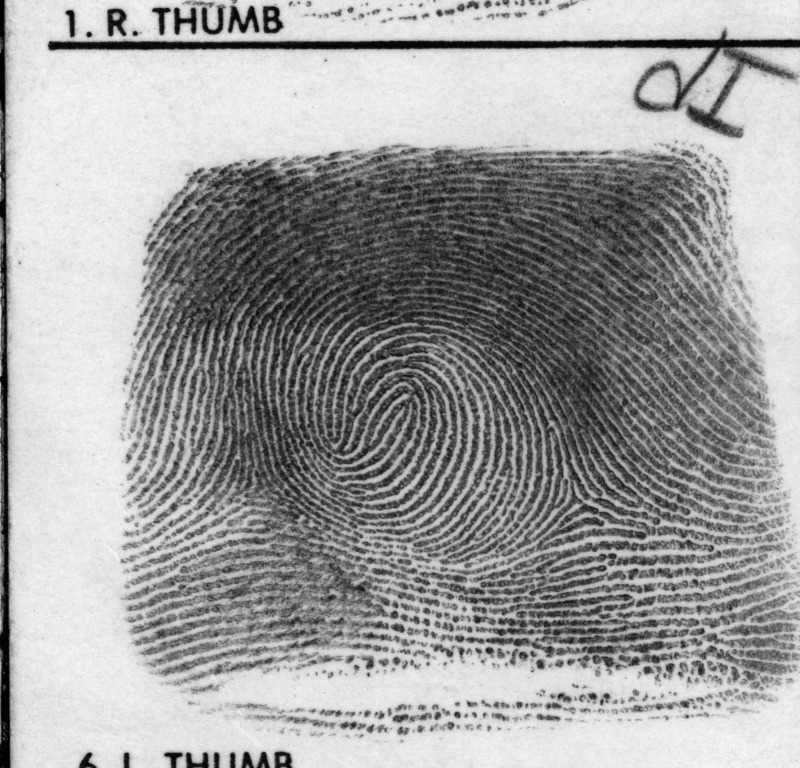}&
            \includegraphics[height=2.4cm]{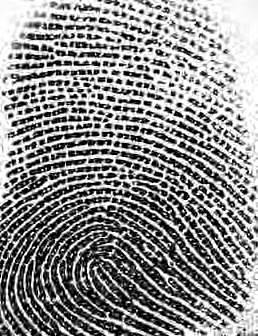}&
            \includegraphics[height=2.4cm]{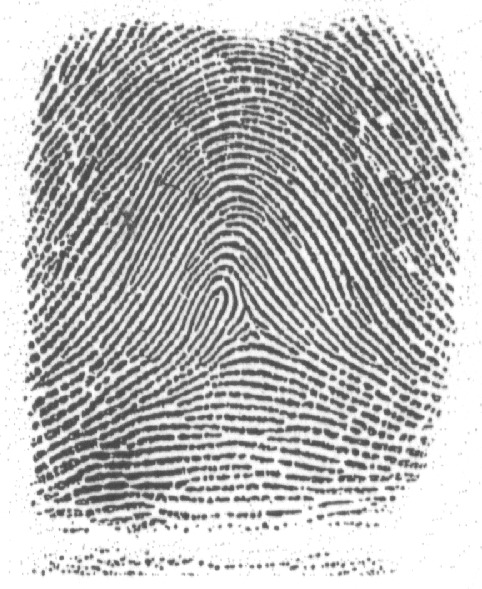}

            \end{tabular}
            \caption{Comparative Display of Latent Fingerprints (top) and Corresponding Rolled Fingerprints (bottom). }\vspace{-0.4cm}
            \label{exampleslatents}
\end{figure}

A notable difficulty in latent fingerprint examination lies in a smaller count of distinct points compared to rolled or slap fingerprints, which are critical for precise pairing. The limited area of the captured fingerprint, often owing to partial or smeared marks, can reduce the number of identifiable distinct points. Moreover, latent fingerprints might contain anomalies, such as deformations in the skin, variations in pressure, or lateral stretching, further complicating the pairing process. A depiction of multiple latent fingerprints and their characteristics, which can complicate matching with rolled fingerprints, is provided in figure \ref{exampleslatents}.

Our \textbf{research objective} is to devise a novel approach for latent fingerprint enhancement centered around a hybrid UNet \cite{ronneberger2015u} architecture. It integrates Resnet-101 \cite{DBLP:journals/corr/HeZRS15}
and UNet encoder functionalities, optimizes ridge and minutiae features using attention mechanisms and skip connections, and incorporates a Gabor layer tailored for GPU computations. Despite being in its initial stages, the ultimate aim is to enhance processing speed, latent fingerprint enhancement adaptability, and feature augmentation. \textbf{Our contributions include:} 
\begin{enumerate}[noitemsep] \item Developing a preliminary approach for latent fingerprint enhancement centered on a hybrid Unet architecture. \item Implementing attention mechanisms and forward skip connections to optimize ridge and minutiae features enhancement in fingerprints. \item A redefined Gabor layer \cite{hong1998fingerprint}, designed for GPU computations, to illustrate the potential acceleration and efficiency gained through modern computational resources. This implementation allows versatile use of this enhancement operation on the GPU, such as CNN or Transformer layer, regarding the magnitude of speed processing. \end{enumerate}

As part of our preliminary research, we have developed an innovative approach to latent fingerprint enhancement, potentially optimizing forensic investigations. While our initial findings indicate the potential to enhance latent fingerprints, \textbf{it is important to note that this is early-stage research, and its application in real-world scenarios should be withheld until further validation and refinement.}

\section{Latent Fingerprint Enhancement Network}

This section presents our proposed model's architecture, explicitly designed to enhance latent fingerprints by closely resembling rolled fingerprints, restoring missing parts, enhancing ridges and minutiae, and reducing noise. The model comprises four main components: an enhanced UNet-like autoencoder, Squeeze-Excitation (SE) blocks, SE attention layers, and a GPU-optimized Fingerprint Gabor Layer. Table \ref{enhancementNew} illustrates the architecture of the latent enhancement process. To train the model, we use the Charbonnier loss function \cite{charbonnier1994two}, which is a robust and differentiable alternative to the L1 loss. It measures the similarity between the output and target images while being less sensitive to outliers. This loss function is well-suited for our model as it ensures the enhancement of latent fingerprints while minimizing the impact of noise and other undesirable artifacts.

\begin{table}[H]
\resizebox{\columnwidth}{!}{%
\begin{tabular}{@{}lc@{}}
\toprule
\multicolumn{1}{l}{\textbf{Layer/Part}} &
  \textbf{\begin{tabular}[c]{@{}c@{}}Channels \\ (Input / Output)\end{tabular}} \\ \midrule
  \textbf{Encoder}    &                                             \\ \midrule
R1. ResNet101 Encoder                  & 1 / 2048    \\
E1. Conv Block \textit{(dilation = 2)} & 1 / 32      \\
E2. Conv Block \textit{(dilation = 2)}          & 32 / 64     \\
E3. Conv Block \textit{(dilation = 2)} & 64 / 128    \\
E4. Conv Block \textit{(dilation = 2)} & 128 / 256   \\
E5. Conv Block \textit{(dilation = 2)} & 256 / 512   \\
CF. Combine Features (E5+R1)           & 2560 / 2560 \\
M. Dense Block                         & 2560 / 1024 \\ \midrule
\textbf{Decoder}    &                                             \\ \midrule
D1. UpConv Block                       & 1024 / 512  \\
D2. UpConv Block + \textit{Skip (E4 → D2) }     & 768 / 256   \\
D3. UpConv Block + \textit{Skip (E3 → D3)}      & 384 / 128   \\
D4. UpConv Block + \textit{Skip (E2 → D4)}      & 192  / 64   \\
F. Conv2d + Sigmoid                    & 64 / 1      \\\midrule
\begin{tabular}[c]{@{}l@{}}G. GPU-Optimized Fingerprint Gabor Layer\\ \end{tabular} &
  1 / 1 \\ \bottomrule
\end{tabular}
}
\caption{Architecture of the solution}

\label{enhancementNew}
\end{table}

\subsection{Enhanced UNet-like Autoencoder}

Our proposed model is based on an enhanced UNet-like autoencoder comprising an encoder, a middle-dense connection layer, and a decoder. The encoder extracts hierarchical features from the latent fingerprint image using convolution blocks for feature extraction and channel-wise recalibration. The decoder then reconstructs the enhanced fingerprint image by employing up-convolution blocks to upscale feature maps and reconstructs the enhanced latent image. The middle-dense connection layer connects the encoder and decoder, integrating dense blocks and an SE Attention Layer to refine the extracted features further. This architecture is essential for restoring missing parts and clarifying ridges and minutiae in latent fingerprints.

Skip connections between encoder and decoder layers maintain spatial information, aiding the reconstruction process. In addition, our implementation incorporates a pre-trained ResNet-101 network as a feature extractor, boosting the model's ability to capture high-level features from the latent fingerprint. 

We also introduce a GPU-optimized Fingerprint Gabor Layer to enhance filtering performance on the reconstructed fingerprint, improving ridge structures and minutiae. This layer utilizes multiple kernel sizes, scales, and orientations to capture complex fingerprint structures more effectively. We explain details of this implementation in the section \ref{gpuoptmizedgaborlayer}.

In summary, the enhanced UNet-like autoencoder uses hierarchical feature extraction with convolution blocks, middle-dense connections with dense blocks, and decoder reconstruction through up-convolution blocks to refine latent fingerprints (see Section \ref{convupconvdenseblocks}). Integrating ResNet-101 for feature extraction and Gabor filtering further optimizes the enhancement process, resulting in more distinct ridge structures and minutiae.

\subsection{Convolution, Up-Convolution, and Dense Blocks}
\label{convupconvdenseblocks}

The convolution, dense, and up-convolutional blocks are essential components of the proposed model, specifically designed to enhance latent fingerprints. The convolution block emphasizes feature extraction and channel-wise recalibration, while the dense block refines extracted features to enhance ridges and minutiae better. The up-convolutional block is responsible for restoring missing parts and improving overall fingerprint quality. These blocks employ a variety of layers and techniques, such as weight-normalized convolutional layers, instance normalization, leaky ReLU activations, dropout layers, and squeeze-excitation layers, all of which contribute to achieving precise fingerprint enhancement. Table \ref{tab:block-components2} summarizing the parts of each block.

\begin{table}[H]
\centering
\footnotesize	
\begin{tabular}{|l|p{4.6cm}|}
\hline
\textbf{Block} & \textbf{Components} \\
\hline
Convolution Block & Weight-normalized Convolutional Layer, Instance Normalization Layer, Leaky ReLU Activation, Dropout Layer, Squeeze-Excitation Layer \\
\hline
Dense Block & Convolutional Layer, Instance Normalization Layer, Leaky ReLU Activation, Self-Attention Mechanism \\
\hline
Up-Convolutional Block & Squeeze-Excitation Layer, Weight-normalized Transpose Convolutional Layer, Instance Normalization Layer, Leaky ReLU Activation, Dropout Layer \\
\hline
\end{tabular}
\caption{Summary of the components of each block on our architecture.}
\label{tab:block-components2}
\end{table}

\subsection{Squeeze-Excitation and SE Attention Layer}
\label{Squeeze}

The Squeeze-Excitation (SE) block \cite{DBLP:journals/corr/abs-1709-01507} enhances latent fingerprints by adaptively recalibrating channel-wise feature responses. This operation explicitly models interdependencies between channels, which is crucial for removing noise and effectively enhancing latent fingerprints. The SE block consists of an adaptive average pooling layer, fully connected layers, activation functions (Leaky ReLU \cite{maas2013rectifier}), and a sigmoid function. By focusing on the most relevant channels, the SE block helps to clarify ridges and minutiae and restore missing fingerprint parts.

The SE Attention Layer extends the SE block by incorporating the self-attention mechanism \cite{DBLP:journals/corr/VaswaniSPUJGKP17} to model long-range dependencies within the input feature map. This layer is essential for capturing critical information in the latent fingerprint, such as ridge structures and minutiae, by considering local and global contexts. It comprises an adaptive average pooling layer, a series of fully connected layers, activation functions (Leaky ReLU), a sigmoid function, and two convolutional layers for query-key-value (QKV) generation and attention computation. The output of the SE Attention Layer is the sum of the input feature map, the SE block output, and the attention output, effectively capturing complex relationships and dependencies within the fingerprint, leading to a more accurate and clearer enhancement.

\subsection{GPU-Optimized Fingerprint Gabor Layer}
\label{gpuoptmizedgaborlayer}

Fingerprint recognition, a key component of biometric identification, presents a considerable computational challenge due to the complex nature of ridge structures and minutiae, the crucial unique identifiers of individual fingerprints. Traditional fingerprint recognition systems' performance depends on the precise extraction and interpretation of these distinctive features. The respected methodology by Lin, Wan, and Jain, while proficient at dealing with rolled and plain fingerprints, faces limitations when processing latent fingerprints. Additionally, its speed constraints and the sequential, handcrafted nature of its processing stages render it ill-suited for integration with deep learning training methods.

In this context, our research proposes an augmented solution that combines the established technique of Lin, Wan, and Jain with modern deep learning methodologies. We have designed a GPU-Optimized Fingerprint Gabor Layer, which expedites the capture and enhancement of latent fingerprint features, rendering it suitable for deep-learning training processes. This layer follows an autoencoder's initial noise reduction and fingerprint reconstruction. Subsequently, the GPU-Optimized Fingerprint Gabor Layer refines the autoencoder output, emphasizing ridges, minutiae, and bridging small occlusions.

Underpinning our development is the utility of Gabor filters, recognized for their efficiency in texture analysis, particularly in fingerprint enhancement. These linear filters emphasize specific frequencies and directions within the spatial domain, illuminating a fingerprint's unique structures and details. Building upon this principle, we crafted a Gabor Layer optimized for GPU execution. A notable attribute of our implementation is its ability to treat convolution operations as parameters learned during the training phase. This dynamic approach transcends the limitations of traditional static kernel-based convolution methods, offering adaptability to diverse fingerprint datasets and creating an adaptive feature extraction system that greatly augments our autoencoder's detail extraction capabilities.

Harnessing the power of GPU parallel processing, our method performs operations in a vectorized and batched manner, leading to a substantial acceleration in processing speed to approximately three orders of magnitude. This efficiency level mirrors other deep learning operations, such as convolutional or Vision Transformer layers, signifying a notable progression in computational speed within biometric processing.

To ensure data consistency, we introduced a contrast stretch technique, normalizing inputs derived from deep layer outputs before processing them by the GPU-optimized Fingerprint Gabor Layer. This normalization ensures uniformity across diverse inputs and enhances the layer's performance by preserving critical data characteristics.

Further bolstering our approach, we have incorporated self-attention mechanisms within the GPU-Optimized Fingerprint Gabor Layer output. These mechanisms selectively concentrate on specific areas necessitating more intricate processing, thereby enhancing information-rich regions. This integration significantly improves the overall effectiveness of our autoencoder system.

The GPU-Optimized Fingerprint Gabor Layer represents an advancement in the field of fingerprint enhancement and can potentially be utilized in diverse scenarios and fingerprint types, not only latents. This refined methodology increases processing speed by integrating traditional fingerprint analysis techniques with the advanced computational capabilities of modern GPUs and self-attention mechanisms. This integrative approach provides a robust foundation for ongoing research in this dynamic field.

\begin{figure*}[htb]
   \centering
     \includegraphics[height=4.0cm]{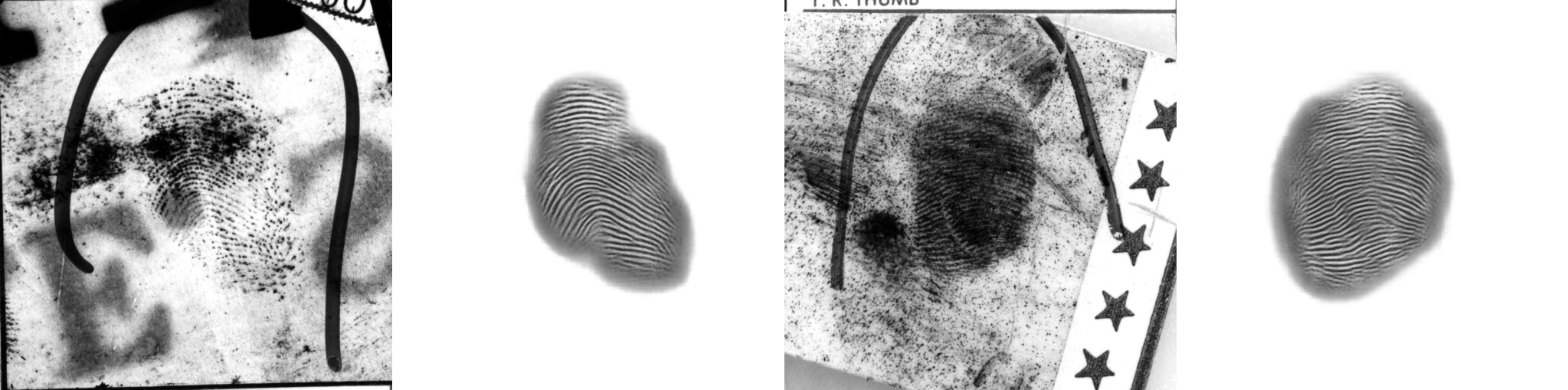}\\
     \includegraphics[height=4.0cm]{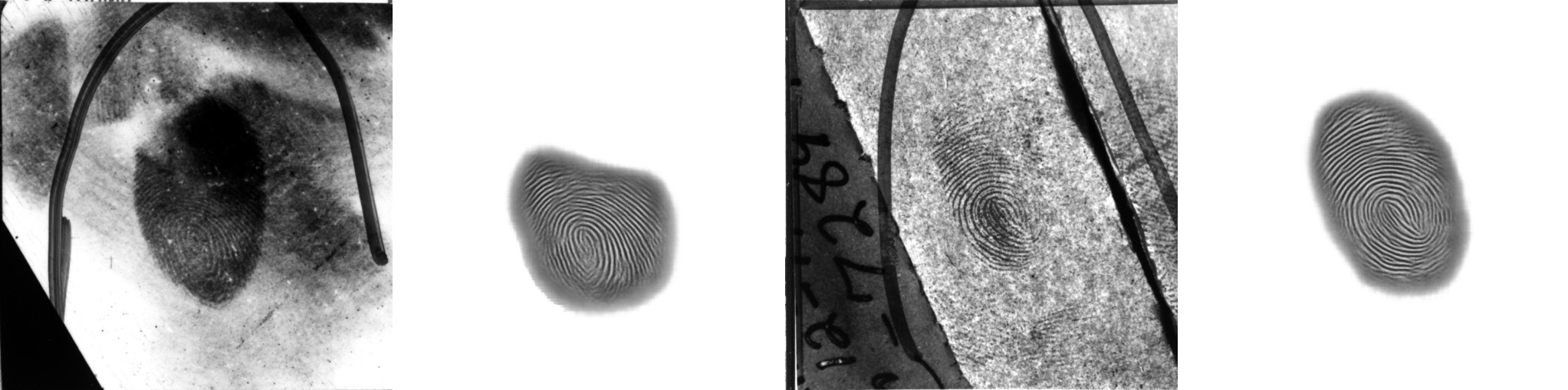}\\
     \includegraphics[height=4.0cm]{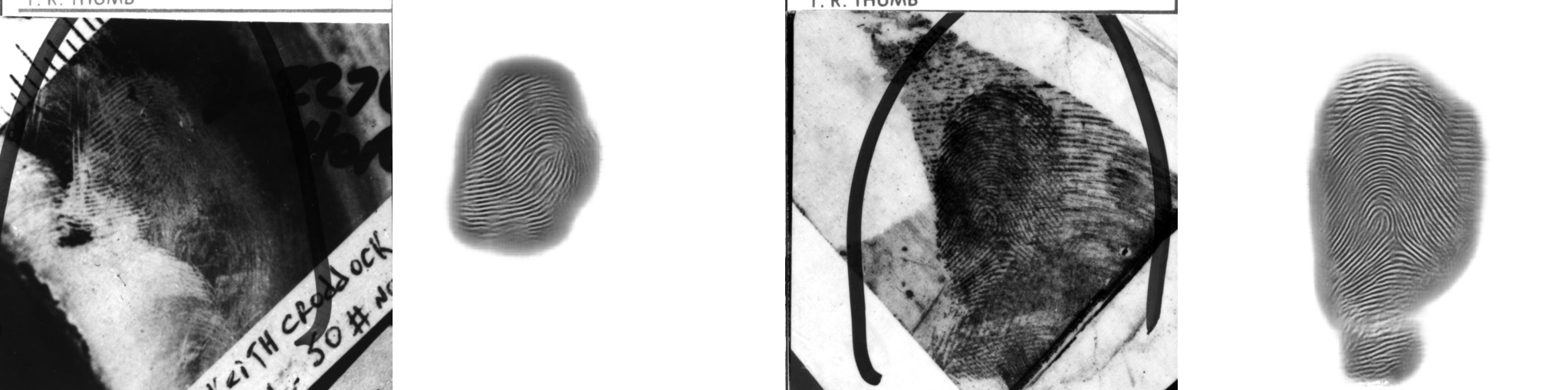}\\
     \includegraphics[height=4.0cm]{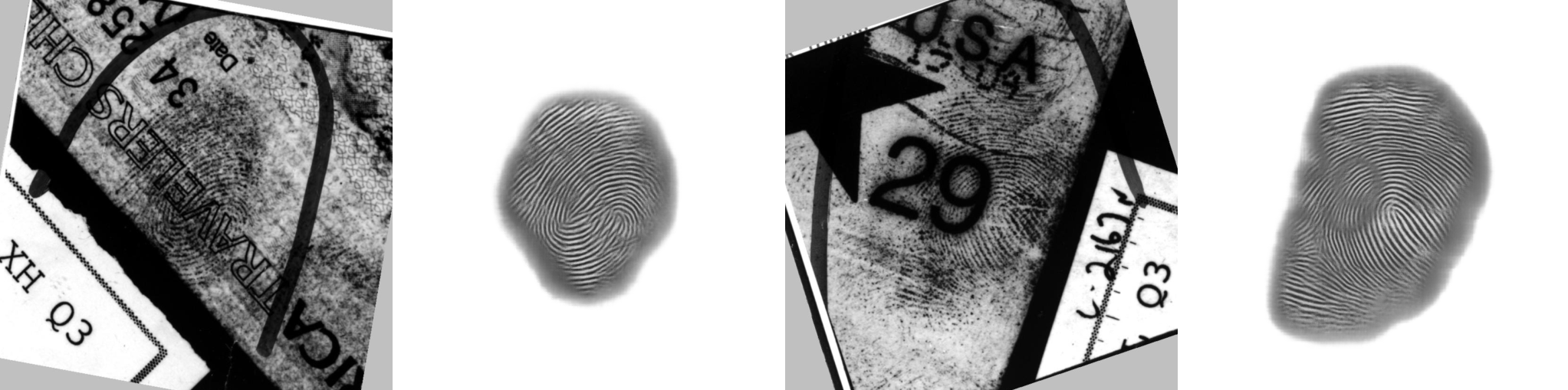}
    \caption{Examples of Latent Fingerprints from NIST SD27 \cite{garris2000nist} and their respective enhanced images.}
    \label{examples}
    \vspace{-0.3cm}
\end{figure*}

\vspace{-0.25cm}
\subsection{Data Augmentation Techniques}

We trained our model using rolled fingerprints from the NIST SD300a database to generate highly realistic latent fingerprint images. To achieve this, we utilized a series of thoughtfully curated data augmentations that mimic the various real-world conditions affecting latent fingerprints. These augmentations provide a diverse and challenging training set.

Firstly, we implemented a texture blending augmentation. This augmentation simulates how fingerprints interact with diverse surfaces by blending the input image with segmented texture images. The process selectively applies textures to image regions, emulating the effect of fingerprints left on textured surfaces like wood, fabric, or metal.

Next, we introduced elastic deformations to simulate fingerprint distortions arising from the skin's elasticity. As a finger presses onto a surface, the skin deforms non-linearly due to its elastic nature, causing the fingerprint to distort. These distortions often appear in real-world scenarios. To replicate this, the augmentation generates a grid of coordinates for the original image and calculates the deformed coordinates by adding the displacement fields. By incorporating such deformations into our training data, our auto-encoder can more accurately restore and enhance latent fingerprints, even in distortions.

Incorporating random noise into the images is another critical augmentation we employ. We simulate real-world imperfections by incorporating varied types of noise, such as Gaussian, salt and pepper, and speckle noise, thus creating a more robust training set. To simulate the fingerprint enhancement techniques used by forensic experts, we apply a dusting powder effect, generating a Perlin noise \cite{perlin1985image} pattern that mimics the uneven distribution of dusting powder commonly seen in practice.

Additionally, we randomly adjust image brightness to simulate variations in lighting conditions. This adjustment helps ensure our recognition system can handle images captured under diverse illumination levels, enhancing its real-world realism and variability. Variations in brightness can occur due to various factors, including different chemical processing techniques, diverse optical camera equipment, and user-applied illumination techniques.

We introduced occlusions or background noise to challenge the model further, such as random letters, linear scratches, free-form lines, ellipses, curves, and rectangles. Including these elements enhances the generalization capabilities of our auto-encoder.

A unique feature of our augmentations is the incorporation of latent fingerprint-smearing effects. This effect simulates the smudging of fingerprints when a finger slides or touches a surface. We utilized Gaussian blur-based smearing, and our approach includes a diverse range of smearing effects found similarly in real-world scenarios. We achieved this by selecting a random kernel size for Gaussian blur, blurring the image, creating an empty mask, and drawing random ellipses on the mask. The result is a partly blurred image that effectively mimics the smearing effect. We then simulate the finger's artificial "dragging" across the surface. Random ellipses from the earlier set are selected, and we use them to create a smear from one random point within the ellipse to another. This process is repeated a random number of times.

Lastly, we implemented an augmentation simulating the impact of different moisture levels on a fingerprint. This augmentation is guided by a parameter ranging from 0, denoting complete dryness, to 1, signifying total wetness. When the moisture level exceeds 0.5, the image undergoes contrast enhancement. This enhancement level is proportional to the moisture level, simulating the dark, pronounced ridges typical of a wet fingerprint. Conversely, if the moisture level falls below 0.5, suggesting drier conditions, we apply brightness enhancement to the image. The degree of this enhancement is inversely proportional to the moisture level, effectively imitating the faded appearance of a dry fingerprint. Moreover, in drier conditions, we enhance the simulation of cracked ridges by incorporating edge details and contours into the image. This is achieved by blending the original image with its edge-detected and contour-filtered versions. The degree of blending depends on the distance of the moisture level from 0.5, hence producing more pronounced cracks under drier conditions.

By combining these augmentations, we have generated a rich and diverse training set of realistic latent fingerprint images. Using our augmented dataset to traditional datasets demonstrates the effectiveness of our approach in creating diverse and challenging latent fingerprint images for training purposes.

\vspace{-0.25cm}

\section{Qualitative results - Visual comparison}

In this section, we showcase the qualitative outcomes of our research through a visual depiction of the enhanced textures obtained using our technique on the NIST SD27 images. This demonstration highlights our approach's potential in generating improved quality latent fingerprint images, emphasizing the preservation of crucial features relevant to fingerprint recognition. It's important to note that this visual comparison does not imply a direct correlation with enhanced performance in fingerprint recognition. Figure \ref{examples} showcases the enhancement achieved by our trained model.

Visually, our model exhibits notable characteristics such as noise reduction, background removal, and enhanced highlighting of the ridges and minutiae in the latent fingerprint images. These improvements can be observed through a clearer representation of the essential details, emphasizing the distinctive patterns and structures important for visual fingerprint analysis.

\vspace{-0.25cm}
\section{Conclusion}

In this work, we have introduced a new approach that merges the robustness of deep learning capabilities with traditional computer vision techniques to augment latent fingerprints. While our approach is in its initial stages, our primary objective is to enhance the processing speed, improve adaptability in latent fingerprint enhancement, and augment fingerprint features. By merging \textit{avant-garde} methodologies from these domains, we aim to improve the conventional boundaries of latent fingerprint enhancement, ultimately leading to a surge in accuracy and reliability of fingerprint recognition systems.

Our approach is based in the hybrid UNet architecture, which is composed by the Resnet-101 network with the UNet encoder. This integration, employing attention mechanisms and forward skip connections, has the objective to enhance the extraction of ridge and minutiae features of latent fingerprints. Moreover, the inclusion of a newly optimized fingerprint Gabor layer, created expressly for GPU operations, underscores the acceleration potential offered by modern computational resources. This versatile layer can operate in a similar way as a CNN or Transformer layer, thus boosting the flexibility and speed of the entire system.

From a visual perspective, our model exhibits enhancements such as noise reduction, background removal, and an amplified emphasis on the ridges and minutiae of latent fingerprint images. These refinements culminate in a clearer representation of important details, thus potentially improving the quality of the images. Future works should consider the following aspects:
\begin{enumerate}[noitemsep]
    \item Full validation through experimental approaches such as open-set (identification 1:N) and open-set validation should be carried out, coupled with the evaluation of fingerprint quality.
    \item The training process can be further fine-tuned by including a minutiae loss function, which would lead to more precise minutiae reconstruction. By incorporating these recommendations, we could potentially refine the accuracy and effectiveness of latent fingerprint enhancements, providing a more reliable tool for fingerprint recognition systems.
\end{enumerate}
\vspace{-0.45cm}



\normalsize
\bibliography{references}


\end{document}